\documentclass[conference]{IEEEtran}
\IEEEoverridecommandlockouts
\usepackage{cite}
\usepackage{amsmath,amssymb,amsfonts}
\usepackage{multicol,multirow}
\usepackage{algorithmic}
\usepackage{graphicx}
\usepackage{textcomp}
\usepackage{xcolor}
\usepackage{colortbl}
\usepackage{xcolor}
\usepackage{booktabs} 
\def\BibTeX{{\rm B\kern-.05em{\sc i\kern-.025em b}\kern-.08em
    T\kern-.1667em\lower.7ex\hbox{E}\kern-.125emX}}
\begin{document}

\title{Versatile and Efficient Medical Image Super-Resolution Via Frequency-Gated Mamba\\
\thanks{* Corresponding author.}
\author{
\IEEEauthorblockN{
Wenfeng Huang\textsuperscript{1,2}, 
Xiangyun Liao\textsuperscript{2}, 
Wei Cao\textsuperscript{3}, 
Wenjing Jia\textsuperscript{1}, and 
Weixin Si\textsuperscript{4*}
}
\IEEEauthorblockA{
\textsuperscript{1}\textit{Faculty of Engineering and Information Technology, University of Technology Sydney, Australia}\\
\textsuperscript{2}\textit{Shenzhen Institutes of Advanced Technology, Chinese Academy of Sciences, China}\\
\textsuperscript{3}\textit{College of Computer Science and Technology, Qingdao University, China}\\
\textsuperscript{4}\textit{Faculty of Computer Science and Control Engineering, Shenzhen University of Advanced Technology, China}\\
}

}}

\maketitle

\begin{abstract}
Medical image super-resolution (SR) is essential for enhancing diagnostic accuracy while reducing acquisition cost and scanning time. However, modeling both long-range anatomical structures and fine-grained frequency details with low computational overhead remains challenging. We propose \textbf{FGMamba}, a novel frequency-aware gated state-space model that unifies global dependency modeling and fine-detail enhancement into a lightweight architecture. Our method introduces two key innovations: a Gated Attention-enhanced State-Space Module (GASM) that integrates efficient state-space modeling with dual-branch spatial and channel attention, and  a \textbf{Pyramid Frequency Fusion Module (PFFM)} that captures high-frequency details across multiple resolutions via FFT-guided fusion. Extensive evaluations across five medical imaging modalities (Ultrasound, OCT, MRI, CT, and Endoscopic) demonstrate that FGMamba achieves superior PSNR/SSIM while maintaining a compact parameter footprint ($<$0.75M), outperforming CNN-based and Transformer-based SOTAs. Our results validate the effectiveness of frequency-aware state-space modeling for scalable and accurate medical image enhancement.

\end{abstract}


\section{Introduction}
High-resolution medical imaging plays a vital role in accurate clinical diagnosis and treatment planning. However, acquiring high-quality images—especially high‑resolution magnetic resonance imaging (MRI)—often requires lengthy scan times and expensive hardware, imposing practical constraints in routine care. Super‑resolution (SR) techniques aim to address this limitation by reconstructing high‑resolution images from low‑resolution acquisitions, thereby enabling detailed anatomical and pathological visualization at reduced cost and time.


The emergence of convolutional neural networks (CNNs) marked a paradigm shift in SR research by enabling end-to-end learning of nonlinear mappings from LR to HR images. SRCNN~\cite{SRCNN} pioneered this approach using a shallow network, which was soon surpassed by deeper and more expressive architectures such as EDSR~\cite{EDSR-baseline}, CARN~\cite{CARN}, and LapSRN~\cite{Lapsrn}, which incorporated residual learning and multi-scale feature fusion. These architectures laid the foundation for a wide array of variants, including frequency-aware designs such as CFSRCNN~\cite{cfsrcnn}, which models coarse-to-fine representations, and LESRCNN~\cite{LESRCNN}, which integrates sub-pixel convolutions with dense blocks for improved efficiency.

In the domain of medical imaging, CNN-based SR models have been widely adopted and extended to accommodate various imaging modalities and clinical requirements. Dual U-Net residual architectures~\cite{qiu2022dual} and volumetric 3D CNNs~\cite{young2024fully} have been leveraged for high-fidelity MRI and cardiac image restoration, preserving anatomical coherence across spatial slices. Generative adversarial networks (GANs) have also proven effective in capturing perceptual realism, as demonstrated in the progressive GAN strategies for MRI and retinal imaging~\cite{mahapatra2019image}. Furthermore, inverse-consistent GANs~\cite{zhang2022frequency} were developed to ensure structural symmetry in OCT super-resolution, while Goyal et al.~\cite{goyal2021weighted} utilized multi-scale cascaded CNNs for ultrasound image enhancement.

To better address frequency-aware signal modeling and modality-specific constraints, recent research introduced attention-based mechanisms and frequency-domain priors. Mix-attention architectures effectively integrate spatial and channel attention in pathological image SR, while some research~\cite{lin2024dual,li2025high} explicitly exploits the Fourier Transform to enhance frequency representation in MRI sequences. These advances reflect the growing recognition of hybrid spatial-frequency modeling as critical for clinical-grade SR performance.

Despite the empirical success of CNN-based techniques, their local convolutional kernels fundamentally limit their ability to model long-distance relationships and global structural patterns. This constraint becomes particularly problematic in high-resolution 3D medical data, where anatomical consistency must be preserved across large spatial extents. 

To overcome the locality limitation of convolutional kernels, Vision Transformers (ViTs) and hybrid CNN–Transformer architectures have been introduced to better model long-distance relationships via global attention mechanisms. Among them, SwinIR~\cite{SWinIR} employs shifted window-based self-attention combined with residual connections and hierarchical representations, striking a balance between global context modeling and computational efficiency. SwinIR and its medical adaptations have shown strong performance across modalities such as MRI and endoscopic imaging, often achieving higher PSNR and SSIM scores than CNN-based counterparts, while also improving perceptual and diagnostic quality~\cite{LGSR}. In particular, SwinIR's ability to capture multi-scale structural priors and semantic coherence has proven beneficial in enhancing subtle anatomical features.

In parallel, other Transformer-inspired models have further enriched the SR landscape. ESRT~\cite{ESRT}, for instance, combines convolutional and Transformer branches through early fusion and skip connections, achieving strong results with relatively low parameter cost. Some researchs~\cite{fang2022cross,li2025high,lin2024dual} combine transformer and frequency information for medical image super-resolution. 

In the medical domain, these models have been increasingly applied across a range of modalities—ultrasound, MRI, CT, endoscopic, and OCT—each with distinct noise characteristics and anatomical priors. To better adapt to these challenges, hybrid frameworks such as LGSR~\cite{LGSR} have emerged. LGSR integrates a local-to-global feature learning pipeline that fuses windowed attention with sparse token selection, enabling efficient contextual interaction while maintaining lightweight design. It demonstrates state-of-the-art performance across ultrasound, OCT, and MRI datasets, offering robust anatomical consistency and high-frequency restoration. 


Recently, structured state‑space models~\cite{mamba,han2024demystify,li2024mamba,muca2024theoretical} (SSMs), and in particular the emergent class of Mamba‑style architectures, have shown remarkable efficiency in modeling long‑range dependencies with linear complexity. Mamba models, offering selective scanning mechanisms and hardware‑aware optimizations, yielding strong sequence modeling power with greatly reduced parameter and memory overheads . The adaptation of Mamba to low‑level vision tasks via MambaIR~\cite{mambair}, which combine convolutional layers and channel attention with state‑space modules to address local spatial recurrences and channel redundancy in restoration tasks. Such researchs~\cite{di2025qmambabsr,xiao2024frequency} has been proven the success of Mamba in image restoration tasks.

Although Transformer-based super-resolution models such as LGSR~\cite{LGSR} have advanced the state of the art in medical image enhancement, their reliance on self-attention mechanisms inevitably introduces quadratic complexity, which limits scalability for high-resolution or volumetric medical data. While CNN-based approaches remain computationally efficient, they struggle to model global dependencies critical for anatomical consistency, especially in modalities such as MRI or OCT where context-aware reconstruction is essential.

To bridge this gap, we propose FGMamba, a novel frequency-aware state-space framework that integrates gated attention and multiscale frequency residual learning for efficient and accurate medical image super-resolution. Inspired by recent advances in state-space sequence modeling, our method captures long-range dependencies with linear complexity while maintaining lightweight design. Unlike prior approaches that rely on windowed or token-based attention, our model introduces a frequency-guided residual feedback mechanism that explicitly enhances high-frequency details—key to restoring structural sharpness in degraded medical scans. Additionally, the gated attention unit selectively enhances discriminative spatial-channel information, further improving texture fidelity and edge continuity.
Our main contributions are summarized as follows:
\begin{itemize}
    
    \item We design a Pyramid Frequency Fusion Module (PFFM) that explicitly enhances high-frequency details by decomposing feature maps across multi-scale FFT domains. This module guides the reconstruction process to recover sharp anatomical boundaries and texture details essential in clinical diagnosis.
    
    \item We introduce a Gated Attention-enhanced State-Space Module (GASM) that augments the vanilla VSSM2D block with spatial and channel attention units. This hybrid design allows selective emphasis on discriminative features while maintaining the memory and runtime efficiency of structured state-space modeling.
    
    \item Extensive experiments across five benchmark medical imaging datasets (ultrasound, OCT, MRI, CT, endoscopic) show that our model outperforms existing CNN-, Transformer-, and Mamba-based SR methods in both PSNR/SSIM and qualitative fidelity—despite using fewer than 0.75M parameters.

\end{itemize}

\begin{figure*}
    \centering
    \includegraphics[width=1\linewidth]{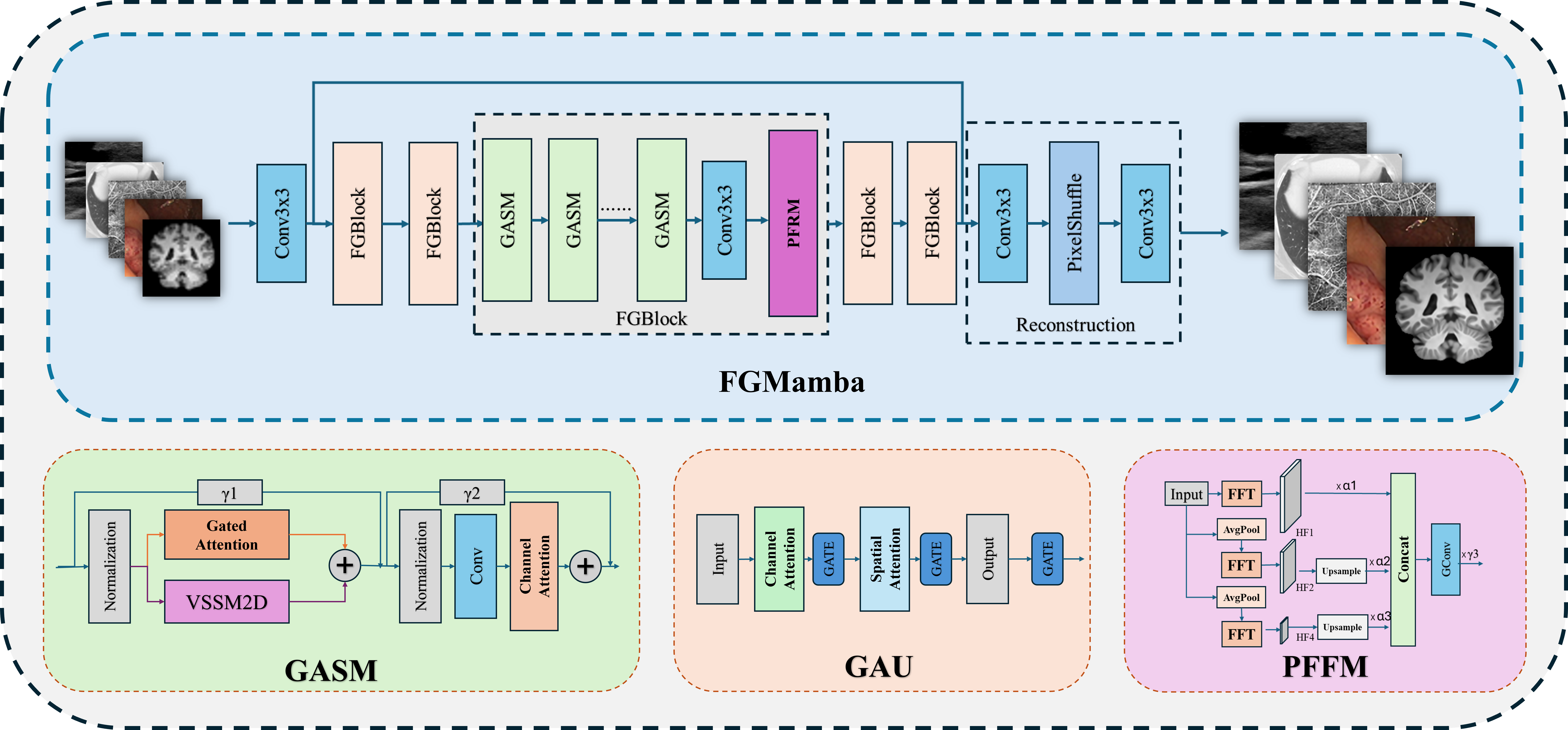}
  \caption{Overall architecture of the proposed \textbf{FGMamba}. It consists of an initial convolution, several FGBlocks, and a reconstruction module with pixel-shuffle upsampling. Each FGBlock contains multiple GASM (Gated Attention State Space Modules), a frequency-enhancing PFFM (Pyramid Frequency Fusion Module), and additional Mamba residual connections. The submodules are illustrated below: (1) GASM incorporates VSSM2D with gated spatial/channel attention, (2) GAU enhances feature selection via dual attention gating, and (3) PFFM extracts and fuses high-frequency components across multiple scales via FFT-based filtering and residual learning.}
  \label{fig:fgmamba_architecture}
\end{figure*}

\definecolor{lightgreen}{RGB}{226,243,234}

\begin{table*}[ht]
\centering
\caption{Quantitative comparison on \textbf{Ultrasound} dataset at $\times$2 scale. Best results are highlighted in \cellcolor{lightgreen}\textbf{bold}.}
\renewcommand{\arraystretch}{1.15}
\setlength{\tabcolsep}{8pt}
\begin{tabular}{l |l |l |c| c c c}
\toprule
\textbf{Method} & \textbf{Architecture} & \textbf{Dataset} & \textbf{Scale} & \textbf{Parameters} & \textbf{PSNR (dB)} & \textbf{SSIM} \\
\midrule
SRCNN~\cite{SRCNN}         & \multirow{5}{*}{CNN}         & \multirow{9}{*}{Ultrasound~\cite{LGSR}} & \multirow{9}{*}{2x} & 69K   & 37.39 & 0.9400 \\
CARN~\cite{CARN}          &                              &                             &                     & 1.59M & 37.68 & 0.9440 \\
EDSR-baseline~\cite{EDSR-baseline} &                              &                             &                     & 1.37M & 37.72 & 0.9447 \\
CFSRCNN~\cite{cfsrcnn}       &                              &                             &                     & 1.49M & 37.62 & 0.9433 \\
LESRCNN~\cite{LESRCNN}       &                              &                             &                     & 0.81M & 37.47 & 0.9428 \\
\cmidrule{1-2} \cmidrule{5-7}
ESRT~\cite{ESRT}          & \multirow{3}{*}{CNN+ViT}     &                             &                     & 0.68M & 37.61 & 0.9436 \\
LBNET~\cite{LBNET}         &                              &                             &                     & 0.73M & 37.51 & 0.9418 \\
LGSR~\cite{LGSR}          &                              &                             &                     & 0.90M & 37.73 & 0.9448 \\
\cmidrule{1-2} \cmidrule{5-7}
\rowcolor{lightgreen}
\textbf{FGMamba} & \textbf{CNN+Mamba}        &                             &                     & \textbf{0.72M} & \textbf{38.13} & \textbf{0.9511} \\
\bottomrule
\end{tabular}
\label{tab:ultrasound_2x}
\end{table*}

\begin{table*}[ht]
\centering
\caption{Quantitative comparison on \textbf{Ultrasound} dataset at $\times$3 scale. Best results are highlighted in \cellcolor{lightgreen}\textbf{bold}.}
\renewcommand{\arraystretch}{1.15}
\setlength{\tabcolsep}{8pt}
\begin{tabular}{l |l |l |c| c c c}
\toprule
\textbf{Method} & \textbf{Architecture} & \textbf{Dataset} & \textbf{Scale} & \textbf{Parameters} & \textbf{PSNR (dB)} & \textbf{SSIM} \\
\midrule
SRCNN~\cite{SRCNN}            & \multirow{5}{*}{CNN}         & \multirow{9}{*}{Ultrasound~\cite{LGSR}} & \multirow{9}{*}{3x} & 69K   & 33.53 & 0.8576 \\
CARN~\cite{CARN}            &                              &                             &                     & 1.59M & 33.66 & 0.8612 \\
EDSR-baseline~\cite{EDSR-baseline}  &                              &                             &                     & 1.55M & 33.71 & 0.8619 \\
CFSRCNN~\cite{cfsrcnn}        &                              &                             &                     & 1.54M & 33.54 & 0.8591 \\
LESRCNN~\cite{LESRCNN}       &                              &                             &                     & 0.81M & 33.56 & 0.8598 \\
\cmidrule{1-2} \cmidrule{5-7}
ESRT~\cite{ESRT}           & \multirow{3}{*}{CNN+ViT}     &                             &                     & 0.77M & 33.66 & 0.8617 \\
LBNET~\cite{LBNET}          &                              &                             &                     & 0.74M & 33.63 & 0.8606 \\
LGSR~\cite{LGSR}            &                              &                             &                     & 0.90M & 33.74 & 0.8622 \\
\cmidrule{1-2} \cmidrule{5-7}
\rowcolor{lightgreen}
\textbf{FGMamba} & \textbf{CNN+Mamba}        &                             &                     & \textbf{0.73M} & \textbf{33.91} & \textbf{0.8659} \\
\bottomrule
\end{tabular}
\label{tab:ultrasound_3x}
\end{table*}

\begin{table*}[ht]
\centering
\caption{Quantitative comparison on \textbf{multi-modal datasets} at $\times$4 scale. Best results are highlighted in \cellcolor{lightgreen}\textbf{bold}.}
\renewcommand{\arraystretch}{1.15}
\setlength{\tabcolsep}{8pt}
\begin{tabular}{l |c|l |c| c c c}
\toprule
\textbf{Method} & \textbf{Architecture} & \textbf{Dataset} & \textbf{Scale} & \textbf{Parameters} & \textbf{PSNR (dB)} & \textbf{SSIM} \\
\midrule
SRCNN~\cite{SRCNN}      &  & OCTA-500~\cite{OCT}  (OCT)         & \multirow{20}{*}{4x} & 69K   & 20.75   & 0.4474 \\
        &                      & CVC-ClinicDB~\cite{endo}(Endoscopic)    &                      &       & 35.65   & 0.9162 \\
        &                      & SARS-COV-2~\cite{CT} (CT)        &                      &       & 33.96   & 0.8645 \\
        &                      & NFBS~\cite{NFBS} (MRI)         &                      &       & 27.76   & 0.8598 \\
\cline{1-1} \cline{3-3}\cline{5-7}
VDSR~\cite{VDSR}    &  \multirow{3}{*}{CNN}                    & OCTA-500~\cite{OCT}  (OCT)         &                      & 0.6M  & 20.86   & 0.4603 \\
        &                      & CVC-ClinicDB~\cite{endo} (Endoscopic)    &                      &       & 36.68   & 0.9269 \\
        &                      & SARS-COV-2~\cite{CT} (CT)        &                      &       & 35.01   & 0.8791 \\
        &                      & NFBS~\cite{NFBS} (MRI)             &                      &       & 29.06   & 0.8943 \\
\cline{1-1} \cline{3-3}\cline{5-7}
Lapsrn~\cite{Lapsrn}  &                      & OCTA-500~\cite{OCT}  (OCT)         &                      & 0.81M & 20.89   & 0.4630 \\
        &                      & CVC-ClinicDB~\cite{endo} (Endoscopic)    &                      &       & 36.33   & 0.9226 \\
        &                      & SARS-COV-2~\cite{CT} (CT)        &                      &       & 35.22   & 0.8815 \\
        &                      & NFBS~\cite{NFBS} (MRI)             &                      &       & 29.58   & 0.9049 \\
\cline{1-2} \cline{3-3}\cline{5-7}
ESRT~\cite{ESRT}     & & OCTA-500~\cite{OCT}  (OCT)     &                      & 0.75M & 20.89   & 0.4627 \\
        &                          & CVC-ClinicDB~\cite{endo} (Endoscopic) &                      &       & 36.12   & 0.9251 \\
        &                          & SARS-COV-2~\cite{CT} (CT)     &                      &       & 35.25   & 0.8815 \\
        &                          & NFBS~\cite{NFBS} (MRI)          &                      &       & 29.38   & 0.9003 \\
\cline{1-1} \cline{3-3}\cline{5-7}
LBNET~\cite{LBNET}    &   \multirow{3}{*}{CNN+ViT}                      & OCTA-500~\cite{OCT}  (OCT)        &                      & 0.74M & 20.90   & 0.4633 \\
        &                        & CVC-ClinicDB~\cite{endo} (Endoscopic)   &                      &       & 36.48   & 0.9238 \\
        &                        & SARS-COV-2~\cite{CT} (CT)       &                      &       & 35.37   & 0.8823 \\
        &                        & NFBS~\cite{NFBS}(MRI)            &                      &       & 29.50   & 0.9031 \\
\cline{1-1} \cline{3-3}\cline{5-7}
LGSR~\cite{LGSR}      &                        & OCTA-500~\cite{OCT}  (OCT)        &                      & 0.9M  & 20.91   & 0.4635 \\
        &                        & CVC-ClinicDB~\cite{endo} (Endoscopic)   &                      &       & 36.88   & 0.9287 \\
        &                        & SARS-COV-2~\cite{CT} (CT)       &                      &       & 35.44   & 0.8840 \\
        &                        & NFBS~\cite{NFBS}(MRI)            &                      &       & 29.75   & 0.9084 \\
\cline{1-2} \cline{3-3} \cline{5-7}
\rowcolor{lightgreen}
\textbf{FGMamba} & \textbf{CNN+Mamba} & OCTA-500~\cite{OCT}  (OCT)     &                      & \textbf{0.74M} & \textbf{20.98} & \textbf{0.4697} \\
\rowcolor{lightgreen}
                 &                    & CVC-ClinicDB~\cite{endo} (Endoscopic)&                      &                 & \textbf{37.32} & \textbf{0.9290} \\
\rowcolor{lightgreen}
                 &                    & SARS-COV-2~\cite{CT} (CT)    &                      &                 & \textbf{36.14} & \textbf{0.8985} \\
\rowcolor{lightgreen}
                 &                    & NFBS~\cite{NFBS}(MRI)         &                      &                 & \textbf{29.91} & \textbf{0.9129} \\
\bottomrule
\end{tabular}
\label{tab:multi_modal_4x}
\end{table*}

\section{Related Works}
\subsection{Image Super-Resolution}
Single-image super-resolution (SISR) has evolved through several phases. Early approaches were grounded in mathematical and statistical techniques such as Random Forest regression, anchored neighborhood regression~\cite{regression}, and dictionary learning, yet they lacked the adaptability to recover complex visual structures. The advent of convolutional neural networks (CNNs) revolutionized SISR: SRCNN~\cite{SRCNN} introduced end-to-end mapping from LR to HR images, and deeper models like EDSR~\cite{EDSR-baseline}, CARN~\cite{CARN}, and LapSRN~\cite{Lapsrn} incorporated residual and pyramid architectures to enhance restoration quality. CFSRCNN~\cite{cfsrcnn} and LESRCNN~\cite{LESRCNN} further emphasized frequency-aware and lightweight design through dense residual connections and sub‑pixel components.

In medical imaging, the adaptation of CNN-based SR has shown strong progress. Dual U‑Net residual~\cite{qiu2022dual} structures and 3D residual CNNs have been used for cardiac MRI restoration, improving local detail preservation and anatomical coherence. GAN-based methods also emerged: Mahapatra et al. employed progressive GANs for retinal and MRI SR~\cite{mahapatra2019image}. OCT super‑resolution~\cite{xu2025axial} has been tried via inverse‑consistent GANs~\cite{zhang2022frequency}, self-supervised learning~\cite{xu2025axial}, and transformer~\cite{yao2024pscat}, and ultrasound SR~\cite{lerendegui2024ultra} has benefited from semi-supervised GAN~\cite{gao2024ultrasound}.  Despite these advances, CNNs remain fundamentally constrained by their local receptive fields, limiting global consistency—especially in large volumetric datasets.

To address this limitation, hybrid CNN‑Transformer architectures were proposed. Such as LGSR~\cite{LGSR}—a CNN‑ViT model designed for medical image SR that combines deformable CNN layers and global Transformers to learn both local detail and long-range semantic context. Such method achieves superior PSNR/SSIM across multiple modalities (Ultrasound, OCT, CT, MRI). However, methods with ViTs still cost a lot because of their self-attention mechanisms.
\subsection{Mamba Architectures}

Mamba~\cite{mamba,han2024demystify} is a recently proposed state-space sequence modeling architecture that offers a compelling alternative to Transformers by addressing their quadratic complexity bottleneck. As a selective state-space model (SSM)~\cite{li2024mamba,muca2024theoretical}, Mamba captures long-range dependencies via continuous-time dynamics while maintaining linear inference complexity, making it highly scalable to high-resolution visual inputs. Unlike traditional RNNs or attention-based mechanisms, Mamba decouples memory access from state updates, allowing dynamic selection of relevant information at each step. This architecture has demonstrated remarkable efficiency in sequence modeling and is now rapidly gaining traction in vision tasks.

The initial success of Mamba in language modeling has spurred several adaptations to low-level vision problems. For example, MambaIR~\cite{mambair} introduces a residual structure that combines spatial encoding and global receptive fields via 2D Mamba modules. These developments~\cite{yuan2025dg,wang2025serp} demonstrate that Mamba-based designs are not only parameter-efficient but also well-suited for tasks requiring both local detail enhancement and global structural understanding. The intrinsic ability of Mamba to model long-term dependencies with low memory overhead makes it an attractive backbone for super-resolution architectures, particularly in the medical domain, where high-resolution volumetric data pose significant computational challenges. 
\begin{figure*}[htpb]
    \centering
    \includegraphics[width=1\linewidth]{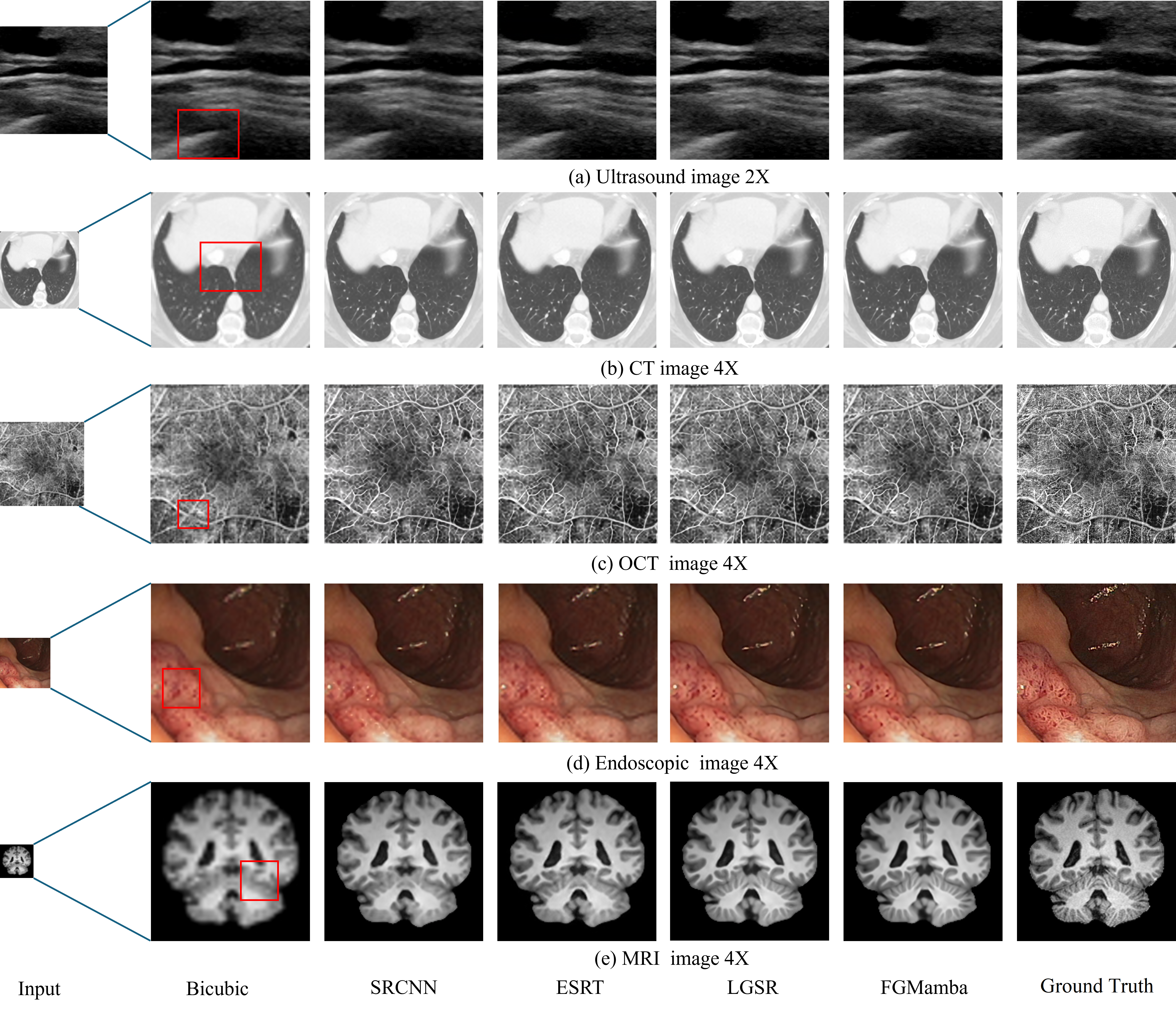}
    \caption{
Visual comparison across five medical modalities: (a) Ultrasound ($\times$2), (b) CT ($\times$4), (c) OCT ($\times$4), (d) Endoscopic ($\times$4), and (e) MRI ($\times$4). Red boxes highlight diagnostically critical structures (vessels, lesions, tissue textures), where FGMamba achieves sharper, more detailed reconstructions to aid radiologist diagnosis and clinical decision-making.
}
    \label{fig:enter-label}
\end{figure*}
By integrating frequency-aware representation and gated attention mechanisms into a lightweight Mamba backbone, our FGMamba architecture enables comprehensive spatial–spectral feature learning while maintaining low parameter overhead. Unlike conventional Transformer or convolutional designs, FGMamba leverages state space modeling for efficient long-range dependency modeling, while simultaneously enhancing fine-grained detail restoration through residual frequency modulation and selective attention. This makes FGMamba a promising solution for high-resolution medical image enhancement across diverse modalities such as OCT, MRI, and CT.

\section{Method}

Our proposed framework, FGMamba, is a compact yet effective medical image super-resolution model inspired the recent researchs~\cite{LGSR,mambair,mamba}. While we inherit the efficient long-range modeling capacity of Mamba, we introduce two key innovations to enhance detail restoration: a Pyramid Frequency Fusion Module (PFFM) and a Gated Attention(GA). The full architecture is illustrated in Fig.~\ref{fig:fgmamba_architecture}.
We begin by applying a $3 \times 3$ convolutional layer to extract shallow features from the low-resolution medical image. Let the input image be denoted as $X$, then the initial feature map is obtained as:
\begin{equation}
F_x = \text{Conv}_{3\times3}(X),
\end{equation}
where $F_x$ represents the extracted shallow feature representation.

And then, we proposed FGBlock, which is the core block of our method.
\subsection{FGBlock}

The FGBlock serves as the core computation module of FGMamba, composed of two essential components: the Gated Attention-enhanced State-Space Module (GASM) and the Pyramid Frequency Fusion Module (PFFM). This dual-branch structure allows the network to effectively combine long-range dependency modeling with fine-grained high-frequency detail enhancement.

\subsubsection*{1) Gated Attention-enhanced State-Space Module (GASM)}

Given the input feature map $F_x \in \mathbb{R}^{C \times H \times W}$, we first apply Layer Normalization:
\begin{equation}
F_{\text{norm}} = \text{Norm}(F_x).
\end{equation}

Then, the normalized feature is processed through two branches:

A state-space branch using the VSSM2D~\cite{mambair} module:
\begin{equation}
F_{\text{vssm}} = \text{VSSM2D}(F_{\text{norm}}),
\end{equation}

and a gated attention branch:
\begin{equation}
F_{\text{gate}} = \text{GA}(F_{\text{norm}}),
\end{equation}
where $\text{GA}(\cdot)$ denotes our proposed Gated Attention Unit and $\text{GATE}(\cdot)$ is a learnable gate controller.
The core of the gated attention branch is our proposed Gated Attention Unit (GAU), which integrates both channel-wise and spatial-wise attention~\cite{Woo} in a lightweight yet effective manner. Specifically, given input $F \in \mathbb{R}^{B \times C \times H \times W}$:

- \textit{Channel Attention}: A global context descriptor is obtained via adaptive average pooling, followed by two fully connected layers to compute a channel-wise attention map $A_c \in \mathbb{R}^{B \times C \times 1 \times 1}$.

- \textit{Spatial Attention}: The input is aggregated across the channel dimension using both average pooling and max pooling. The concatenated result is passed through a $k \times k$ convolution (default $k{=}7$)  to yield a spatial attention map $A_s \in \mathbb{R}^{B \times 1 \times H \times W}$.

The final attended feature is computed as:
\begin{equation}
F_{\text{gate}} = F \odot (A_c\odot Gate \odot A_s \odot Gate) \odot Gate,
\end{equation}
where $\odot$ denotes element-wise multiplication. And the $Gates$ are learned by $Sigmoids$. This gating mechanism enables the network to selectively emphasize semantically informative features while suppressing irrelevant background noise.

The outputs from both branches are fused with a learnable residual scaling parameter $\gamma_1$:
\begin{equation}
F_{\text{add}} = F_{\text{vssm}} + F_{\text{gate}} + \gamma_1 \cdot F_x.
\end{equation}

Then, we pass $F_{\text{add}}$ through a convolution and channel attention block:
\begin{align}
F_{\text{norm2}} &= \text{Norm}(F_{\text{add}}), \\
F_{\text{conv}} &= \text{Conv}_{3\times3}(F_{\text{norm2}}), \\
F_{\text{gasm}} &= \text{Conv}_{3\times3}(\text{Channel Attention}(F_{\text{conv}}) + \gamma_2 \cdot F_x).
\end{align}

\subsubsection*{2) Pyramid Frequency Fusion Module (PFFM)}

To extract texture-rich information, we introduce a frequency-domain enhancement module. For each scale $s \in \{1, 2, 4\}$, the input $F_{\text{gasm}}$ is downsampled:
\begin{equation}
F_s = 
\begin{cases}
F_{\text{gasm}}, & \text{if } s = 1, \\
\text{AvgPool}(F_{\text{gasm}}, s), & \text{otherwise}.
\end{cases}
\end{equation}

The 2D Fourier Transform is applied:
\begin{equation}
\mathcal{F}_s = \text{FFT}(F_s),
\end{equation}
followed by a high-frequency mask:
\begin{equation}
M_s = \mathbb{1}(|\mathcal{F}_s| > \mu_{\mathcal{F}_s}),
\end{equation}
and inverse FFT to get high-frequency spatial features:
\begin{equation}
H_s = \text{IFFT}(\mathcal{F}_s \cdot M_s).
\end{equation}

Each $H_s$ is upsampled back to original resolution and weighted by a learnable coefficient $\alpha_s$:
\begin{equation}
F_s^{\uparrow} = \alpha_s \cdot \text{Upsample}(H_s).
\end{equation}

The fused high-frequency feature is:
\begin{equation}
F_{\text{PFFM}} = \gamma \cdot \text{GroupConv}_{1\times1}\left( \text{Concat}(F_1^{\uparrow}, F_2^{\uparrow}, F_4^{\uparrow}) \right),
\end{equation}
where $\gamma$ is a learnable scale parameter. The feature representation after passing through multiple FGBlocks is denoted as $F_{\text{FGB}}$.

\subsubsection*{4) Reconstruction Module}

After passing through a series of FGBlocks, the final feature map is denoted as $F_{\text{final} = F_(FGB) + F_x}$. To recover the high-resolution image from this deep representation, we adopt a simple yet effective reconstruction pipeline.

Specifically, the reconstruction block consists of a $3 \times 3$ convolution  to refine the features, followed by a PixelShuffle operation~\cite{pixel} to upscale the spatial resolution. A final $3 \times 3$ convolution layer is then applied to generate the output image:
\begin{equation}
I_{\text{SR}} = \text{Conv}_{3\times3}\left(\text{PixelShuffle}\left(\text{Conv}_{3\times3}(F_{\text{final}})\right)\right),
\end{equation}
where $I_{\text{SR}}$ denotes the super-resolution image.

\section{Experiments}
\subsection{Datasets}
\subsubsection{Ultrasound Image Dataset}
 We use the breast ultrasound dataset~\cite{LGSR}. It includes 500 high-resolution scans acquired using GE Vivid Iq and E9 systems in ``Breast'' mode. After sliding-window cropping and augmentation, we generate 12,000 training and 1,250 testing patches of sizes $240\times240$ and $256\times256$, respectively. LR images are created via bicubic downsampling at scales $2\times$, and $3\times$.

\subsubsection{OCT Image Dataset}
We use the ``OCTA-6M-Projection Map-OCTA(FULL)'' subset from OCTA-500~\cite{OCT}, which contains 300 retinal images ($400\times400$ pixels). For SR tasks, we downsample them to $100\times100$ via bicubic interpolation and split the dataset into 80\% for training and 10\% each for validation and testing.

\subsubsection{Endoscope Image Dataset}
Colonoscopy frames are selected from CVC-ClinicDB~\cite{endo}, consisting of 612 images extracted from 31 video sequences. To remove irrelevant dark and overlaid regions, images are cropped to $240\times240$ and downsampled to $60\times60$ for paired training. We adopt an 8:1:1 split for training, validation, and testing.

\subsubsection{CT Image Dataset}
We utilize the SARS-CoV-2 CT dataset~\cite{CT}, selecting the 1,230 scans from healthy individuals. After cropping via a sliding window, we obtain 5,432 CT patches at $200\times200$ resolution. LR versions are generated by bicubic downsampling. The dataset is split into 983 training, 123 validation, and 123 testing samples.

\subsubsection{MRI Image Dataset}
MRI data is sourced from the NFBS repository~\cite{NFBS}, which provides 125 skull-stripped T1 brain scans. We extract the final 30 slices per scan—regions with fuller anatomical content—and crop each to $160\times160$. LR images ($40\times40$) are created by downsampling. The split includes 3,000 training, 375 validation, and 375 testing images.

\subsection{Impletement details}

Following standard practices in recent literature~\cite{LGSR,mambair}, we augment the training data using horizontal mirroring and random rotations of $90^\circ$, $180^\circ$, and $270^\circ$. For patch-based learning, we divide each image into fixed-size patches, with patch dimensions dynamically adjusted depending on the dataset and upscaling factor. To maintain consistency across experiments, we adopt a batch size of 8. Optimization is conducted using the Adam algorithm with momentum parameters set to $\beta_1 = 0.9$ and $\beta_2 = 0.999$. The learning rate is initialized at $2 \times 10^{-4}$. All experiments are carried out on a single NVIDIA RTX 4090 GPU.
\begin{table}[t]
\centering
\caption{Ablation studies on OCT (CT) dataset at $\times$4 scale.}

\setlength{\tabcolsep}{10pt}
\begin{tabular}{lcc}
\toprule
\textbf{Method}     & \textbf{PSNR (dB)} & \textbf{SSIM} \\
\midrule
Baseline (without GAU \& PFFM) & 20.9750           & 0.4686       \\
w/o GAT (no GAU in GASM)  & 20.9765           & 0.4693       \\
w/o Freq (no PFFM frequency module)     & 20.9740           & 0.4684       \\
\rowcolor{lightgreen}
\textbf{FGMamba (full model)} & \textbf{20.9781}           & \textbf{0.4697}       \\
\bottomrule
\end{tabular}
\label{tab:ablation}
\end{table}
\subsection{Comparison with the State of the Arts}

To evaluate the effectiveness of our proposed FGMamba we conduct comprehensive comparisons against a range of state-of-the-art (SOTA) super-resolution models, including traditional CNN-based methods (SRCNN~\cite{SRCNN}, VDSR~\cite{VDSR}, LapSRN~\cite{Lapsrn}), Transformer-integrated variants (ESRT~\cite{ESRT}, LBNET~\cite{LBNET}), and recent gated attention approaches (LGSR~\cite{LGSR}). Quantitative results across multiple medical image modalities—Ultrasound, CT, OCT, endoscopic, and MRI—are presented in Tables~\ref{tab:ultrasound_3x}, \ref{tab:ultrasound_3x}, and \ref{tab:multi_modal_4x}. 

As shown, FGMamba consistently achieves superior performance with significantly fewer parameters. For instance, under the challenging \textbf{$\times4$} scale across multi-modal datasets (Table~\ref{tab:multi_modal_4x}), FGMamba attains the highest PSNR/SSIM scores in all domains, such as 37.32 dB / 0.9290 (endoscopic), 36.14 dB / 0.8985 (CT), and 29.91 dB / 0.9129 (MRI), outperforming the ViT-based LBNET and LGSR models while using fewer parameters (0.74M vs. 0.9M). Similarly, for \textbf{$\times3$} scale on ultrasound images (Table~\ref{tab:ultrasound_3x}), FGMamba leads with 33.91 dB / 0.8659, demonstrating its robustness in lower-resolution recovery tasks. 

The qualitative results shown in Fig.~\ref{fig:enter-label} further highlight FGMamba’s ability to restore high-frequency textures and preserve anatomical structures across different imaging types. Compared to other methods, FGMamba produces visually sharper boundaries, more realistic textures, and fewer artifacts, especially evident in vascular (OCT) and gastrointestinal (endoscopic) scenes. These improvements not only enhance perceptual fidelity but also provide clearer visualization in diagnostically critical regions, facilitating more accurate lesion detection and anatomical analysis. 


\subsection{Ablation Study}
To validate the effectiveness of our proposed components, we conduct an ablation study on the OCT dataset at $\times 4$ scale. As shown in Table~\ref{tab:ablation}, removing the gated attention mechanism in GASM slightly degrades both PSNR and SSIM, confirming its contribution to structure enhancement. Similarly, excluding the frequency-domain PFFM module results in further performance drops, indicating the importance of multi-scale high-frequency restoration. The full FGMamba model achieves the best results, demonstrating the complementary benefits of both modules.
\section{Discussion}
As shown in Fig.~\ref{fig:enter-label}, the red box regions highlight diagnostically important structures such as vessels, lesions, and tissue textures. FGMamba produces sharper and more detailed reconstructions in these areas, which can assist radiologists in improved diagnosis and clinical decision-making. Specifically, our frequency-aware gated state-space architecture generates significantly sharper anatomical boundaries and richer textural details compared to existing SOTAs. Gated Attention-enhanced State-Space Module synergizes dual-branch spatial and channel attention with efficient state-space modeling, while the Pyramid Frequency Fusion Module exploits FFT-guided fusion to captures high-frequency details across multiple resolutions. Clinically, these enhancements directly translate to improved diagnostic confidence in identifying early-stage pathologies and refining treatment planning. Furthermore, the enhanced perceptual quality across five modalities (Ultrasound, OCT, MRI, CT, Endoscopy) facilitates downstream tasks like segmentation or detection, potentially boosting the accuracy of automated analysis. Given its lightweight design ($<$0.75M parameters) and modality-agnostic framework, FGMamba shows strong potential for deployment in clinical systems. In future work, we plan to evaluate its impact on representative downstream tasks to further validate its practical utility.

\section{Conclusion}

In this paper, we introduced FGMamba, a lightweight and frequency-aware super-resolution framework tailored for medical imaging. By integrating a gated attention mechanism with structured state-space modeling (GASM) and enhancing high-frequency detail via a pyramid frequency fusion module (PFFM), our method effectively captures both global contextual patterns and fine structural cues. FGMamba demonstrates superior PSNR and SSIM performance across five distinct medical modalities while maintaining under 0.75M parameters. Extensive evaluations validate its ability to recover sharp anatomical boundaries and texture details, offering a promising and scalable solution for clinical image enhancement tasks. Future work may explore its extension to volumetric SR and real-time deployment in diagnostic systems.

\bibliographystyle{IEEEtran}
\bibliography{ref}
\end{document}